\documentclass[letterpaper, 10 pt, conference]{IEEEtran}

\IEEEoverridecommandlockouts
\usepackage[top=72pt, bottom=57pt, left=54pt, right=54pt]{geometry}
\usepackage[utf8]{inputenc}
\usepackage[pdftex]{graphicx}
\usepackage{float}
\usepackage{amsmath}
\usepackage{amsthm}
\usepackage{circuitikz}
\usepackage{caption}
\usepackage{fancyhdr}
\usepackage{verbatim}
\usepackage{subcaption}
\usepackage{ dsfont }
\title{Folding Assembly}
\usepackage{url}
\usepackage{mathtools}
\usepackage{flushend}
\newcommand{\mtr}{\ensuremath{^\top}}

\mathtoolsset{showonlyrefs}

\newtheoremstyle{example}
  {6pt}
  {6pt}
  { }
  {}
  {\bfseries}
  { }
  { }
  { }
\theoremstyle{example}
\newtheorem{example}{Example}
\newtheoremstyle{problem}
  {6pt}
  {6pt}
  {\itshape}
  {}
  {\bfseries}
  { }
  { }
  { }
\theoremstyle{problem}
\newtheorem{problem}{Objective}
\newtheorem{remark}{Remark}

\begin{document}
%
\title{\LARGE{\bf{Folding Assembly by Means of Dual-Arm Robotic Manipulation}}}
\author{
	\IEEEauthorblockN{Diogo Almeida}
	\and
	\IEEEauthorblockN{Yiannis Karayiannidis}
	\thanks{The authors are with the Computer Vision and Active Perception Lab., Centre for Autonomous Systems, School of Computer Science and Communication, Royal Institute of Technology KTH, SE-100 44 Stockholm, Sweden.
e-mail: \tt{ $\{$diogoa$|$yiankar$\}$@kth.se}}
  \thanks{Y. Karayiannidis is with the Dept. of Signals and Systems, Chalmers University of Technology, SE-412 96 Gothenburg, Sweden, e-mail: \tt{ yiannis@chalmers.se}}
  \thanks{This work has been carried out in the SARAFun project, partially funded by the EU within H2020 (H2020-ICT-2014/H2020-ICT-2014-1) under grant agreement no. 644938}
}

\maketitle

\begin{abstract}
In this paper, we consider \textit{folding assembly} as an assembly primitive suitable for dual-arm robotic assembly, that can be integrated in a higher level assembly strategy.
The system composed by two pieces in contact is modelled as an articulated object, connected by a prismatic-revolute joint.
Different grasping scenarios were considered in order to model the system, and a simple controller based on feedback linearisation is proposed, using force torque measurements to compute the contact point kinematics.
The folding assembly controller has been experimentally tested with two sample parts, in order to showcase folding assembly as a viable assembly primitive.
\end{abstract}

\section{Introduction}
Robotic assembly is one of the main tasks performed by robots in industrial settings, but has only been applied in a subset of potential cases, as for example in the automotive industry.
In typical industrial settings stationary single-arm robots are employed to perform assembly tasks that consists of assembly primitives within a higher level logic, taking full advantage of the structured industrial workspace.
%

Recently, there is an increasing trend of employing dual-arm human-sized robots to perform a variety of tasks that are typically executed by humans, in dynamically changing environments originally designed for human use \cite{Kock2011}.
Cooperation with humans has also been a key motivation for using dual-arm robots.
For instance, the inherent anthropomorphism of a robot with two arms makes its movements more readily predictable when observed by a human \cite{Park2008, Nakai2006}.
Furthermore, dual-arm manipulators can enrich the set of assembly primitives and thus widen the range of the assembly tasks that can be performed by a robot, allowing for an assembly execution that is independent of environmental fixtures.
The later is crucial for environments where uncertainty is a major factor, such as human-centered industrial environments.

%

In this work, we introduce \textit{folding assembly}, an assembly primitive that specifically takes into account the rotational motion of the parts, while maintaining contact between both pieces.
In particular, the assembly task is accomplished by firstly bringing the parts into contact and subsequently gradually adjusting their relative poses.
We perform a kineto-statics analysis that focuses on different contact settings and robotic grasps, and present a velocity control strategy that allows its successful execution.
The controller is based on feedback linearisation of the contact point kinematics and thus its performance depends on the calculation of the contact using force/torque measurements.
Experimental evaluation demonstrates a folding assembly task and the effect of inaccurate calculation of the contact point.

The remaining of this manuscript is structured as follows.
In section \ref{relatedWork} we provide background on work in robotic manipulation.
A description of the folding assembly problem is present in section \ref{problemDescription}.
A kineto-statics analysis is done in section \ref{kinetoStaticAnalysis}, and a control strategy to regulate the task is derived in section \ref{controllerDerivation}, with the respective experimental results shown in section \ref{results}.
Finally, in section \ref{conclusion}, conclusions and objectives for future work are discussed.
\begin{figure}
	\centering
	\includegraphics[width = 0.3\textwidth]{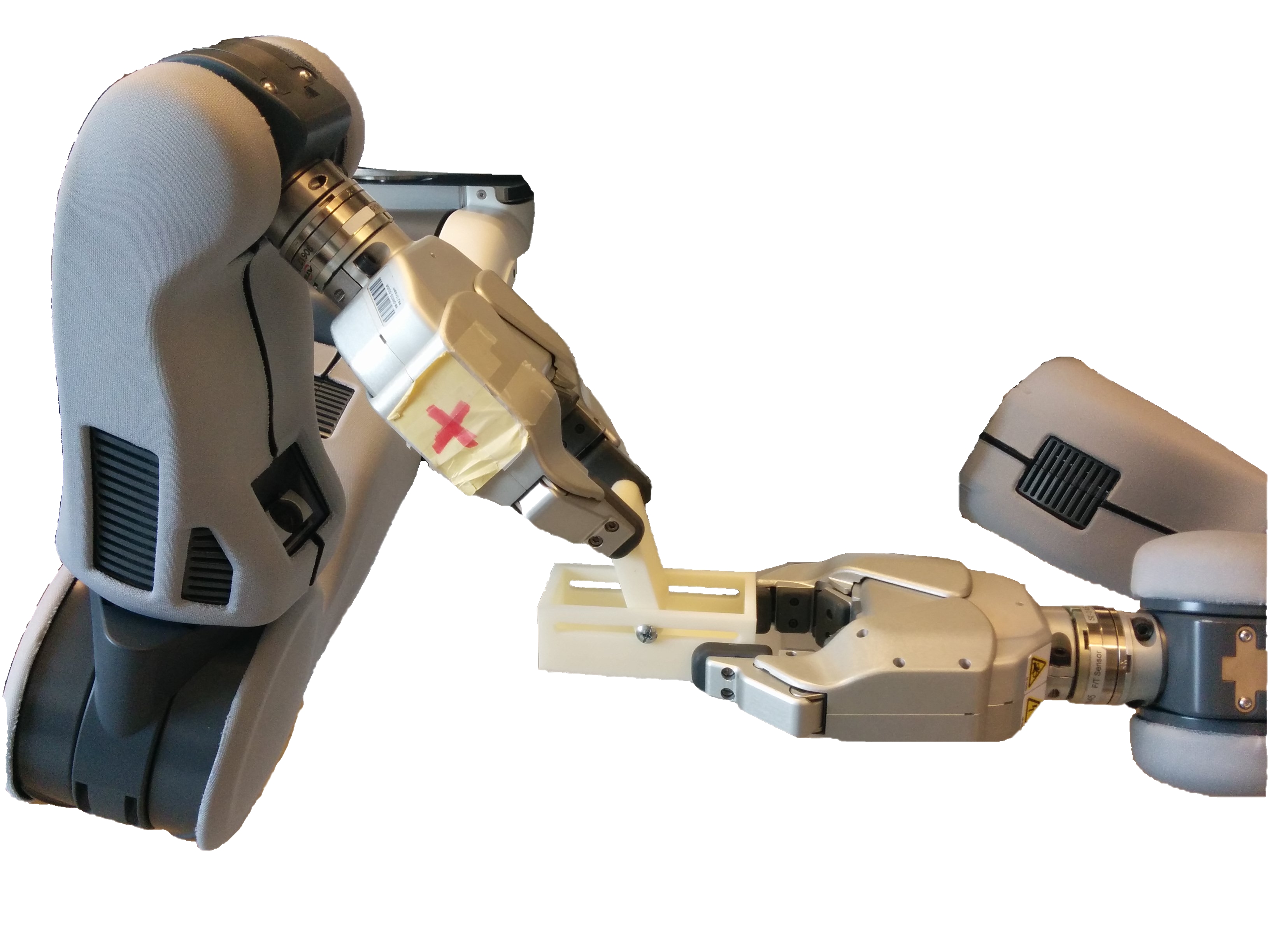}
	\caption{Folding assembly\label{roboticSchematic}}
\end{figure}

\section{Related work\label{relatedWork}}
Robotics applications are related to a wide variety of use cases.
In industrial robotic applications, precision is a mandatory requirement which can be fulfilled given the low degree of uncertainty in traditional industrial settings. To overcome unavoidable imprecisions in path planning algorithms, force feedback was integrated in the control strategy design.
This is important in contact tasks, such as assembly operations, where forces need to be regulated simultaneously with the  assembly motion generation \cite{Lefebvre2005, MccarragherBrenanJ1993}. This led to the development of impedance \cite{Kazerooni1986} and hybrid force position controllers \cite{Khatib1987}, where the manipulator objective remains to achieve a desired motion, while regulating forces along the contact surfaces.

More recently, there has been a push towards getting robots to work in unstructured environments.
Moving the robots away from the traditional industrial settings increases the necessity of designing systems that can safely and robustly operate under a much larger amount of uncertainty. 
This can be achieved by the construction of robots that are intrinsically safe \cite{Kock2011} and the development of robot control strategies that aim at preventing damage from unexpected collisions \cite{Vick2013, Flacco2012}. Furthermore, the use of robots in human-centered environments has motivated the design of dual-arm manipulators that can be designed to be anthropomorphic and behave in a human-like way \cite{Nakai2006, Albers2006}.

\begin{figure}
	\centering
	\begin{subfigure}[b]{0.1\textwidth}
		\includegraphics[width = \textwidth]{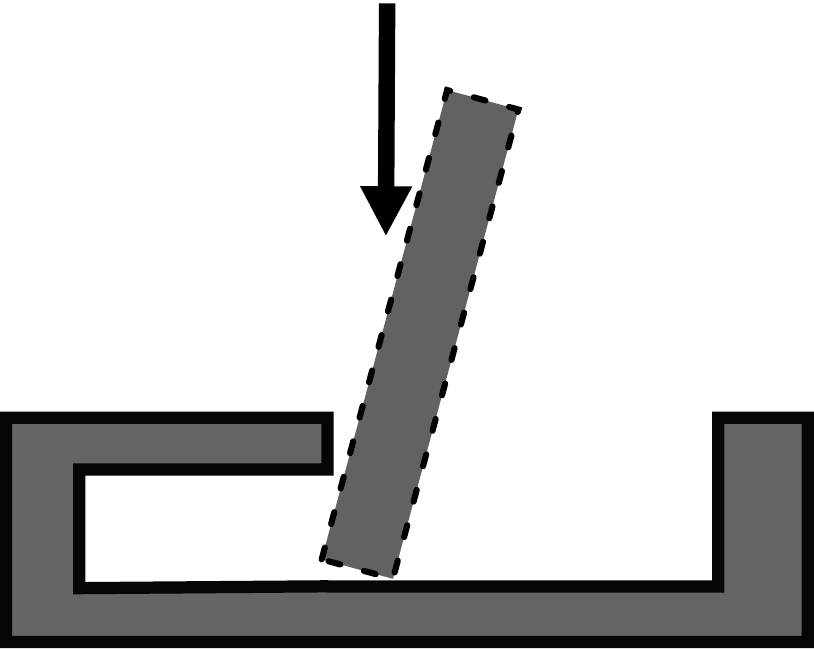}
		\caption{}
	\end{subfigure}
	\qquad
	\begin{subfigure}[b]{0.1\textwidth}
		\includegraphics[width = \textwidth]{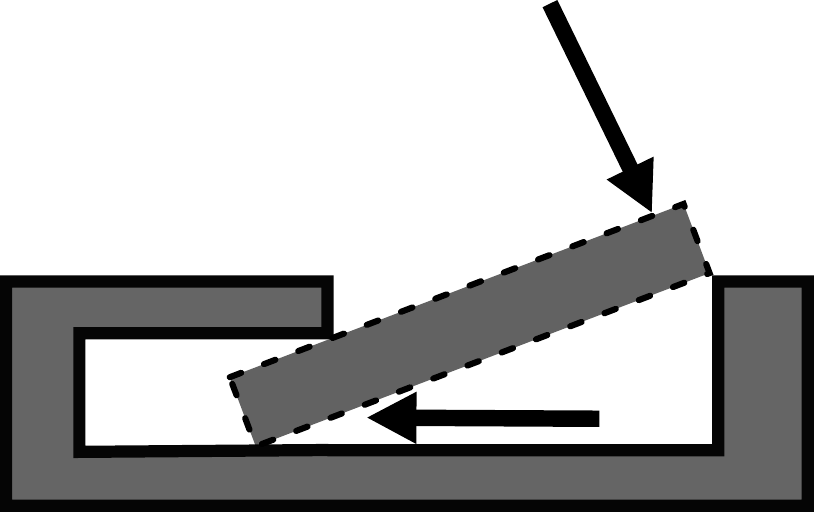}
		\caption{}
	\end{subfigure}
	\qquad
	\begin{subfigure}[b]{0.1\textwidth}
		\includegraphics[width = \textwidth]{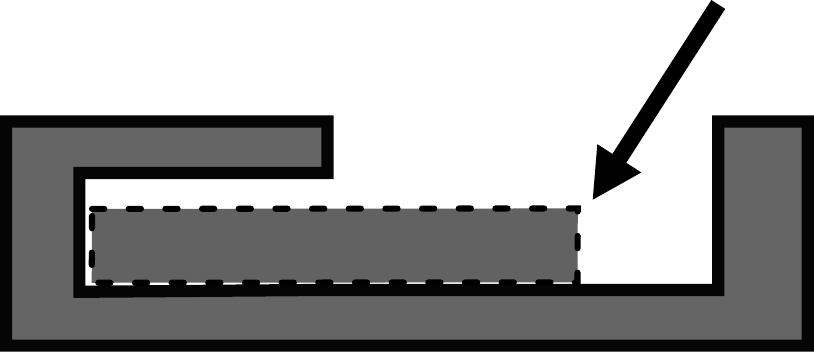}
		\caption{}
	\end{subfigure}
	\caption{Folding assembly scenario requiring the object to fit into a tight socket. \label{socketFold}}
\end{figure}

 Dual-arm robots can enhance human-robot interaction for augmenting human capabilities \cite{Kruger2009} and the way a robot is programmed can be simplified via the introduction of high level semantics \cite{Surdilovic2010}.
In particular, in \cite{Makris2014}, a programming library was developed that allows for a user to program dual-arm robot tasks visually.
This is done by using vision and voice recognition to interpret the desired functionality (move down,  rotate, approach, etc.).

An effective way to create modular and complex assembly systems is to define a set of assembly primitives, and adopt a higher level logic that switches between them in order to obtain a desired assembly behaviour.
For instance, in \cite{MccarragherBrenanJ1993}, petri nets are used to model transitions between different discrete steps of an assembly problem while an automatic assembly planner was proposed in \cite{Finkemeyer2005}, where each assembly task is executed by a suitable controller; switching between states is made when a set of limit conditions is reached.
More recently, \cite{Wang2014} employs the concept of function blocks for robotic assembly.
Each primitive assembly task is encapsulated in a block, every one of which with its own set of techniques for the successful assembly execution.
A complete assembly task is divided in these primitives, and petri nets are used to model the transitions between different assembly operations.

Robotic assembly with dual-arm manipulators has advantages and poses specific challenges; an in-depth review of the state-of-the-art for dual arm manipulation can be found in \cite{Smith2012}.
The increased flexibility and larger independence from a structured environment comes at the cost of an increase in system complexity.
In particular, interaction forces between the manipulators and grasped objects require careful consideration, due to the closed kinematic chain \cite{Yun1991}. While both single and dual arm robots can be used for applying planning strategies \cite{MccarragherBrenanJ1993, Finkemeyer2005, Wang2014}, a dual arm robot can take advantage from the cooperation between manipulators that allows for a reduction in task complexity \cite{Kruger2011}.
Several works detail master-slave implementations.
When carrying a load, it is common to have one arm perform the motion control of the overall system, while its slave uses force feedback to ensure tracking of the object changing position \cite{Mukaiyama1996, Caccavale2001}.
Master-slave relationships are also relevant in scenarios where the need to overcome a physical distance between an operator and its work subject is present \cite{Tavakoli2003, Carignan2004}.

Our contribution with this manuscript is twofold:
\begin{itemize}
	\item Add to the pool of assembly primitives by modelling a generic \emph{folding} motion, where one part is expected to slide and rotate against the other;
	\item Implement folding assembly in a dual-arm robot, manipulating two distinct objects by maintaining an unilateral contact constraint.
	In order to achieve this, a master-slave relationship between arms is employed.
\end{itemize}

\section{Problem description\label{problemDescription}}
In this section, we provide a high level description of folding assembly, introduce some key concepts to be used in the remaining sections and provide a couple of illustrating examples.
\subsection{Surface and rod piece}
Consider two pieces that will be assembled together and are grasped by a distinct robotic manipulator so that each piece can be manipulated independently.
The assembly task is then carried out in a two-step approach.
Firstly, we achieve contact between pieces in an appropriate spot.
Secondly, we adjust this contact point/edge position, and reorient the pieces, until they are assembled as intended.

 \begin{figure}
	\centering
	\begin{subfigure}[b]{0.15\textwidth}
		\includegraphics[width = \textwidth]{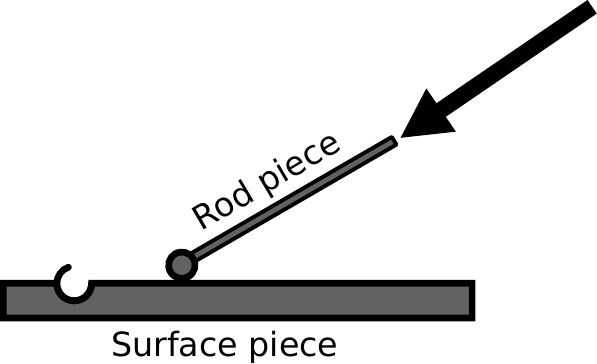}
		\caption{}
	\end{subfigure}
	\qquad
	\begin{subfigure}[b]{0.15\textwidth}
		\includegraphics[width = \textwidth]{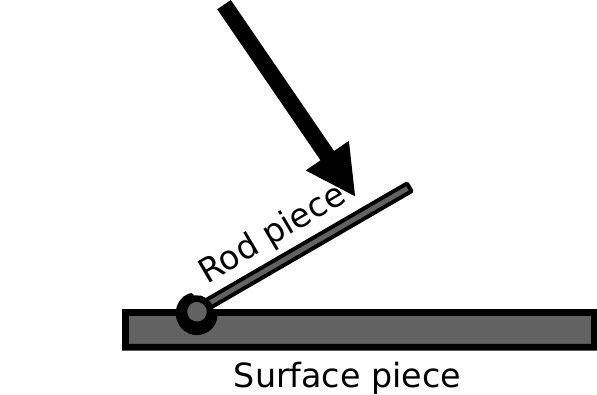}
		\caption{\label{lidRevolute}}
	\end{subfigure}
	\caption{Assembly by prior insertion on a socket. \label{socketHinge}}
\end{figure}

We name the two pieces as the \textit{surface piece} and the \textit{rod piece} (fig. \ref{socketHinge}), joined together via a sliding contact point:
\begin{itemize}
	\item The \textit{surface piece} is the part along which the contact point translates. It should usually be the part that will host the other one after the assembly. In a single-arm scenario, it would be the part held by environmental fixtures;
	\item Conversely, the \textit{rod piece} is the part that will translate along with the contact point, and would be the grasped part in the single-arm case;
	\item The motion of the independent pieces can be represented as that of an articulated object with two rigid bodies connected via a prismatic-revolute joint. Sliding and rotating the parts corresponds to a change in the joint state, as long as contact is maintained.
\end{itemize}

Using a virtual joint to model the contact between parts is helpful in visualizing the overall conceptualized motion.
Note, however, that non-ideal effects, such as stick-slip friction, moving the rod around its friction cone, or inserting it in a surface piece fixture, leads to switches between different modes of operation for the virtual joint. 
In particular, when friction prevents sliding of the contact point the virtual joint operate as a purely revolute joint.

	\subsection{Examples}
	Consider the following examples:
	
	\begin{example}
		Battery insertion in a cellphone
		
		\emph{Using one end-effector to grasp the device, the battery compartment cover lid must be removed and dropped in an appropriate place. The battery can then be grasped and inserted in the battery socket.
		The robot grasps the lid again and folds it back in position.}\label{batteryExample}
	\end{example}
	
	\begin{example}
		Closing a box cover
		
		\emph{By aligning its edges with a box, the cover should be moved along it, until the box cover fixtures are reached.
		The cover can then be rotated unto the box, and snapped into position.}\label{lidExample}
	\end{example}
	
	The aforementioned tasks can be modelled as sequences of folding assembly steps.
	Consider example \ref{batteryExample}.
	Initially, removing the cover lid is essentially doing folding disassembly.
	It is required to maintain a force along the lid, while performing the opening motion, until the lid is in position to be removed.
	Battery insertion can be done by establishing contact close to the connectors on the socket edge. It can then be folded into position by rotating the battery while sliding the contact edge against the extremity.
	In both cases, the cellphone acts as the surface piece, while	 the lid and battery are the two consecutive rod pieces.
	
	Similarly in example \ref{lidExample} the rod piece can be readily identified as being the lid, with the container as the surface piece.
	By placing one of the lid edges on the container, it can then slide until it matches the corresponding container edge (as if on rails).
	The lid can then be rotated to complete the task.
	
	Folding can be the only feasible solution for some problems.
	For instance, if we consider the battery compartment of example \ref{batteryExample} to be as depicted in fig. \ref{socketFold}, it is not possible to directly insert the battery, and a combined sliding rotating movement needs to be undertaken.
	Example \ref{lidExample} can be solved by carefully placing the lid along the edges of the container, and then pressing.
	In cases where it becomes unfeasible to grasp the lid by holding two parallel edges with the same gripper, it may be convenient to perform the assembly as in picture \ref{socketHinge}.	

\subsection{Dual-arm folding assembly}
The folding task can be analysed by taking into consideration both the way pieces are being grasped and the semantic roles of each assembly part.
Two main strategies for the assembly execution can be considered:

\begin{itemize}
	
	\item \textbf{Master-Slave}:
		Distinct roles can be assigned to the manipulators, making one perform the overall motion, while the other acts almost as an environmental fixture and extra sensor;
	\item \textbf{Cooperative}:
		Both manipulators can perform independent motion, in order to achieve the assembly goal.
		Separate objectives can be considered in this case as well.
		For instance, the surface piece end-effector may adjust its part to better ensure contact conditions.
\end{itemize}

Humans approach this kind of assembly tasks differently for different scenarios.
Removing a cellphone cover lid as in example \ref{batteryExample} can be done in different ways depending on the part design.
If there is a slot where we can insert our finger, applying pressure in the direction of the hinge and rotating is enough to open the compartment.
This do not require grasping of the piece, but just the presence of a contact point.
Meanwhile, the main body of the phone is rigidly grasped by the other hand.

In case of a human grasping both parts, the assembly activity is rarely executed with a perfectly rigid grasp.
Being able to perform independent rotations between the end-effector and the grasped part allows for executing the task successfully, while minimizing the overall motion of the arms.
We can define two major ways of grasping each part: a) \textbf{Rigid grasp}: where the grasped piece cannot independently rotate with respect to the end-effector and b) \textbf{Non-rigid grasp}: or soft grasp, where the grasped piece is allowed to rotate with respect to the end-effector, around the grasping point.

In the following section, we will perform the kineto-statics analysis for the folding assembly, assuming the ideal case of an articulated object with a prismatic-revolute joint.

\section{Kineto-statics analysis\label{kinetoStaticAnalysis}}
We begin by defining the notation and formal problem description to be used in the modelling of the folding assembly problem.
From the kinematics of the system, we derive the expected reaction forces for different grasping scenarios.

\subsection{Notation}
\begin{itemize}
	\item Lower case letters in bold denote vectors, while bold upper case letters will be used for matrices. The $\cdot^\top$ symbol is used for the transpose of a vector or matrix.
	\item The pose of a frame $\{i\}$ with respect to a frame $\{j\}$ is described by the position vector $^j\mathbf{p}_i \in \mathds{R}^m$ and a rotation matrix $^j\mathbf{R}_i\in SO(m)$. We will consider the spatial case $m=3$. The superscript will be omitted when referring to the inertial frame $\{\mathbf{O}\}$.
	\item The matrix $\mathbf{S}(\boldsymbol \omega)$ is the skew-symmetric matrix,
	\begin{equation}
		\mathbf{S(}\boldsymbol \omega) = \begin{bmatrix}
			0 & -\omega_z & \omega_y\\
			\omega_z & 0 & -\omega_x\\
			-\omega_y & \omega_x & 0
		\end{bmatrix},
	\end{equation}
	that performs the cross product between $\boldsymbol \omega$ and a vector $\mathbf{x} \in \mathds{R}^3$, that is, $\boldsymbol \omega \times \mathbf{x} = \mathbf{S}(\boldsymbol \omega) \mathbf{x}$.
	Note that $\dot{\mathbf{R}} = \mathbf{S}(\boldsymbol \omega)\mathbf{R}$, where $\boldsymbol \omega$ is the angular velocity of the frame with orientation $\mathbf{R}$.
\end{itemize}

\subsection{Description of the assembly problem}
\begin{figure}
	\centering
	\includegraphics[width=0.35\textwidth]{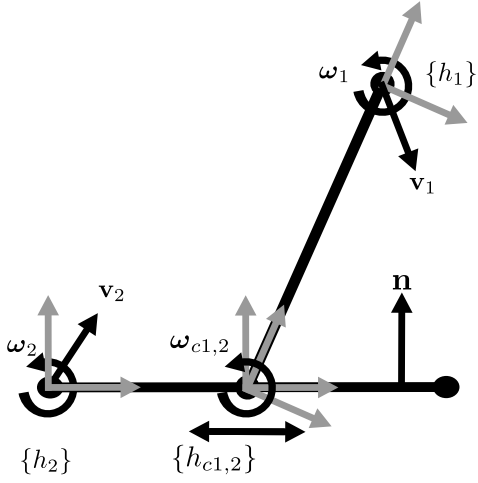}
	\caption{Folding assembly model. Grey arrows illustrate the coordinate frames $\{h_i\}$\label{foldingModel}}
\end{figure}

Let the two robotic end-effectors be described by the indices $j \in \{1,2\}$, for the hand grasping the rod and the surface piece respectively. We define $\{h_1\}$ and $\{h_2\}$, attached to the  grasping point of the rod piece and the surface piece respectively; the orientation of the frames is related to the orientation of the end-effectors. Let $\mathbf{p}_c$ denote the contact point and $\{h_{c1}\}$, $\{h_{c2}\}$ denote frames that are attached to the contact point with orientations related to the orientation of the rod and surface piece respectively.
We denote the position and orientation of the frames with respect to $\{\mathbf{O}\}$, with $\mathbf{p}_i$ and $\mathbf{R}_i$, $i \in \{1,2,c1,c2\}$ respectively. We also define the vectors $\mathbf{r}_j$, connecting the two grasping points with the contact point as follows:
\begin{equation}
	\mathbf{r}_j = \mathbf{p}_c - \mathbf{p}_j, \quad j\in\{1,2\}.
	\label{rVectors}
\end{equation}
While the vector $\mathbf{r}_1$ related to the rod piece has fixed length, denoted by $l$, i.e. $\|\mathbf{r}_1\| = l$, vector $\mathbf{r}_2$ has a varying length during the assembly execution that represents the translation variable of the virtual joint (prismatic joint). Assuming fixed orientation for $\{h_{c2}\}$, $\mathbf{R}_{c1}$ can be utilised for representing the rotational variable of the virtual joint (revolute joint).

The linear and angular velocities of frame $i$ are denoted by $\mathbf{v}_i = \dot{\mathbf{p}}_i$ and $\boldsymbol \omega_i$. By differentiating (\ref{rVectors}) and $\mathbf{r}_1 = \mathbf{R}_{c1}\, ^{c1}\mathbf{r}_{1}$ with respect to time and taking into account  $^{c1}\dot{\mathbf{r}}_1 = 0$ we can derive the following constraint:
\begin{equation}
	\mathbf{v}_1 = \mathbf{v}_c - \mathbf{S}(\boldsymbol \omega_{c1})\mathbf{r}_1.
	\label{velocityExpression}
\end{equation}
that relates the velocity of the end-effector grasping the rod-piece $\mathbf{v}_1$ with the contact point velocity $\mathbf{v}_c$ and the rotational velocity of the rod $\boldsymbol \omega_{c1}$.
A similar relation can be obtained for the surface piece end-effector, by expressing $\mathbf{v}_2$ in terms of components tangential and normal to the surface piece,
\begin{equation}
		\mathbf{v}_2 = \mathbf{v}_{2\parallel} - \mathbf{S}(\boldsymbol \omega_{c2})\mathbf{r}_{2}.
	\label{surfacePieceVelocity}
\end{equation}
Furthermore, in case of rigid grasps the relative orientation of an end-effector and the corresponding piece is invariant and the rotational velocities of frames $\{h_1\}$, $\{h_2\}$ and $\{h_{c1}\}$, $\{h_{c2}\}$ are subjected to the following constraints:
\begin{equation}
	\boldsymbol \omega_j = \boldsymbol \omega_{cj}, \quad j\in\{1,2\}.
	\label{rigidGraspConstraint}
\end{equation}

%
%
%
%
%
%
%

Taking into account the constraints on the velocities, we can derive the expected reaction wrench due to contact between the pieces that are reciprocal to the robot movement directions \cite{DeLuca1994}, $\mathbf{w}_{R_i} = [\mathbf{f}_i\;\boldsymbol \tau_i]^\top$, where $\mathbf{f}_i$ and $\boldsymbol \tau_i$ are respectively the force and torque components of the reaction wrench. 
We get:

\begin{equation}
	[\mathbf{v}_i\mtr \;\boldsymbol \omega_i \mtr ]\,\mathbf{w}_{R_i} = 0.
	\label{reactionForce}
\end{equation}


 We will consider the cases of a contact point which can slide along the surface and a fixed contact point which can allow only rotations of the rod piece. In both cases we consider a rigid grasp for the surface piece while rigid and non-rigid grasp subcases are considered for the end-effector grasping the rod piece.

\subsection{Sliding contact point}
In case of a \textbf{non-rigid grasp}, the rotation of the end-effector is unconstrained i.e. Eq. \eqref{rigidGraspConstraint} does not hold and there is only one configuration that allows for the rod grasping end-effector to exert a force to the surface piece. In particular when the rod piece is normal to the surface, the end-effector velocity becomes constrained along the surface normal, i.e. $\mathbf{n}^\top\mathbf{v}_1 = 0$
%
and hence  \eqref{reactionForce} implies that $\mathbf{w}_R$ can be defined as follows:
\begin{equation}
	\mathbf{w}_{R_1} = \begin{bmatrix}
		\mathbf{n}\\
		0
	\end{bmatrix} \lambda_{R_1},
	\label{forceRodNonRigidSliding}
\end{equation}
with a $\lambda_{R_1}$ that can be considered a Lagrange multiplier measured in Newtons (N). In case of a \textbf{rigid grasp}, both \eqref{velocityExpression} and \eqref{rigidGraspConstraint} hold, and since the contact point is constrained to move along the surface tangent i.e. $\mathbf{n}^\top \mathbf{v}_c = 0$, the following constraint can be derived:
\begin{equation}
		\begin{bmatrix}
			\mathbf{n}^\top & -\mathbf{n}^\top \mathbf{S}(\mathbf{r}_1)
		\end{bmatrix} \begin{bmatrix}
			\mathbf{v}_1\\
			\boldsymbol \omega_1
		\end{bmatrix} = 0.
	\label{rodRigidSlidingConstraints}
\end{equation}
Hence, both forces and torques can be exerted in this case:
\begin{equation}
	\mathbf{w}_{R_1} = \begin{bmatrix}
		\mathbf{n}\\
		\mathbf{S(r}_1) \mathbf{n}
	\end{bmatrix}\lambda_{R_1}.
	\label{forceRodRigidSliding}
\end{equation}

Regarding the surface piece end-effector, we can derive the expected reaction forces by performing the inner product of the surface normal with \eqref{surfacePieceVelocity} and in turn utilising \eqref{reactionForce}. Since a rigid grasp is only considered for the surface piece the result is similar to \eqref{forceRodRigidSliding}:
\begin{equation}
	\mathbf{w}_{R_2} =
	\begin{bmatrix}
		\mathbf{n}\\
		\mathbf{S(r}_2)\mathbf{n}
	\end{bmatrix}\lambda_{R_2}
	\label{surfaceSlidingRigid}
\end{equation}

\subsection{Fixed contact point}
A fixed contact point implies the following constraint for the contact point velocity:
\begin{equation}\label{fixed_velocityzero}
	\mathbf{v}_c = 0.
\end{equation}
In case of a \textbf{non rigid grasp}, the translational velocity of the rod piece end-effector is constrained along the radial direction of the motion and thus reaction forces can arise along $\mathbf{r}_1$, i.e:
\begin{equation}
	\mathbf{w}_{R_1} = \begin{bmatrix}
		\displaystyle \frac{\mathbf{r}_1}{\|\mathbf{r}_1\|}\\
		0
	\end{bmatrix} \lambda_{R_1},
\end{equation}
Note that no torque can be applied by the non-rigid grasp about the axis of free rotation.
In case of a \textbf{rigid grasp}, the constraint (\ref{velocityExpression}) for the rod piece end-effector can be written as follows:
\begin{equation}
	\mathbf{v}_1 - \mathbf{S(r}_1) \boldsymbol \omega_1 = 0.
	\label{rodPieceConstrainedVelocity}
\end{equation}
while the corresponding reaction forces satisfying (\ref{reactionForce}) can be defined as follows:
 \begin{equation}
	\mathbf{w}_{R_1} = \begin{bmatrix}
		\mathbf{I}\\
		-\mathbf{S(r}_1)
	\end{bmatrix} \boldsymbol\lambda_{R_1},
	\label{rodPieceConstrainedForces}
\end{equation}
Note that the fixed contact point allows for forces to be applied along the rod independently of its orientation while a rigid grasp allows for force to be applied along all spatial directions.

Regarding the surface piece end-effector, any movement along the surface piece will be constrained, $\mathbf{v}_{2\parallel} = 0$, and as such \eqref{surfacePieceVelocity} becomes equivalent to \eqref{rodPieceConstrainedVelocity}. Hence the reaction wrench will have the same form as in \eqref{rodPieceConstrainedForces}:
\begin{equation}
	\mathbf{w}_{R_2} = \begin{bmatrix}
		\mathbf{I}\\
		-\mathbf{S(r}_2)
	\end{bmatrix} \boldsymbol\lambda_{R_2}.
	\label{forcesStaticContact}
\end{equation}

%

\section{Control of a folding assembly task\label{controllerDerivation}}
\begin{figure}
	\centering
	\includegraphics[width = 0.3 \textwidth]{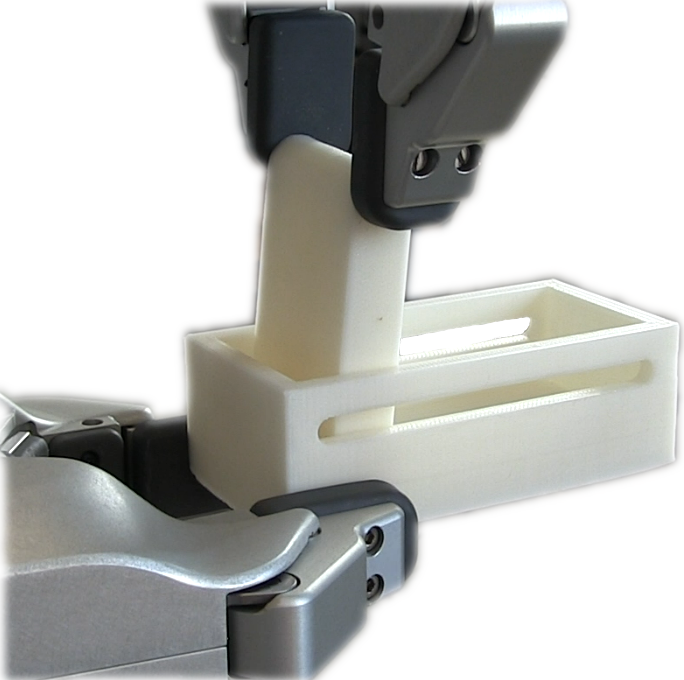}
	\caption{Folding assembly parts\label{foldingObject}}
\end{figure}

We consider that the assembly parts are rigidly grasped at each point, and adopt a master-slave relationship between the end-effectors.
The surface piece end-effector remains static during the assembly, $\mathbf{v}_2 = \mathbf{0}$ m/s and  $\boldsymbol \omega_2 = \mathbf{0}$ rad/s, while the rod piece end-effector performs the overall motion. The contact point kinematic model is given by:
\begin{equation}
	\begin{array}{l}
		\mathbf{v}_c = \mathbf{S}(\boldsymbol \omega_1)(\mathbf{p}_c - \mathbf{p}_1) + 		\mathbf{v}_1\\
		\boldsymbol \omega_c = \boldsymbol \omega_1
	\end{array}.
	\label{openVC}
\end{equation}
and the following  control objectives are set:
\begin{problem}
	{\normalfont\fontfamily{cmr}\selectfont Velocity tracking}
	
	Make the contact point move with the velocity profile $[\mathbf{v}_{\text{d}},\; \boldsymbol \omega_{\text{d}}]$.
	\label{velocityObjective}
\end{problem}
\begin{problem}
	{\normalfont\fontfamily{cmr}\selectfont Contact enforcement}

	Keep the reaction forces close to a desired value, $[\mathbf{f}_d,\;\boldsymbol \tau_d]$, ensuring that contact is kept.
	\label{forceObjective}
\end{problem}

Assuming that the inner velocity control loop of the manipulator allows us to command with minor errors the end-effector velocity, we will design a set of commands $\mathbf{v}_1,\;\boldsymbol \omega_1$ in order to achieve objectives \ref{velocityObjective} and \ref{forceObjective}.
The velocity controllers are designed in to feedback linearise \eqref{openVC} and to command reference linear  and angular velocities denoted by $\mathbf{v}_{\text{ref}}$ and $\boldsymbol \omega_{\text{ref}}$  as follows:
\begin{equation}
	\begin{array}{l}
		\mathbf{v}_1 = -\mathbf{S}(\boldsymbol \omega_1)(\mathbf{p}_c - \mathbf{p}_1) + \mathbf{v}_{\text{ref}}\\
		\boldsymbol \omega_1 = \boldsymbol \omega_{\text{ref}}
	\end{array},
	\label{controlEquation}
\end{equation}
with
\begin{align}
\mathbf{v}_{\text{ref}}& = \mathbf{v}_d + \mathbf{v}_f  \quad \text{where} \quad \mathbf{v}_f = -K_f f_e \mathbf{n},\\
\boldsymbol{\omega}_{\text{ref}}& = \boldsymbol{\omega}_d
\end{align}
where $\mathbf{v}_d$ is the desired linear velocity profile for the contact point along the surface piece tangent, $\boldsymbol{\omega}_d$ is the desired angular velocity of the rod piece, $f_e$ is the error between the normal contact force magnitude and the desired one, and $K_f$ is the force control gain.
The control input $\mathbf{v}_f$ is employed in order to achieve objective \ref{forceObjective} while $\mathbf{v}_d$ and $\boldsymbol{\omega}_d$ can be either designed as motion control inputs utilising feedback of the rod piece pose error or feedforward terms that remain active until a higher control loop detects a change in the assembly state.
\begin{figure}
	\centering
	\begin{subfigure}[b]{0.13\textwidth}
		\includegraphics[width = \textwidth]{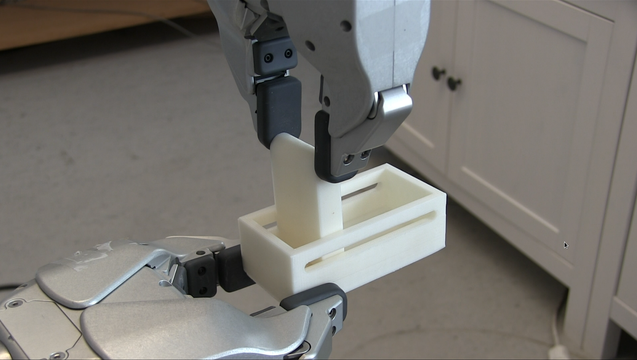}
		\caption{\label{vid1}}
	\end{subfigure}
	\begin{subfigure}[b]{0.13\textwidth}
		\includegraphics[width = \textwidth]{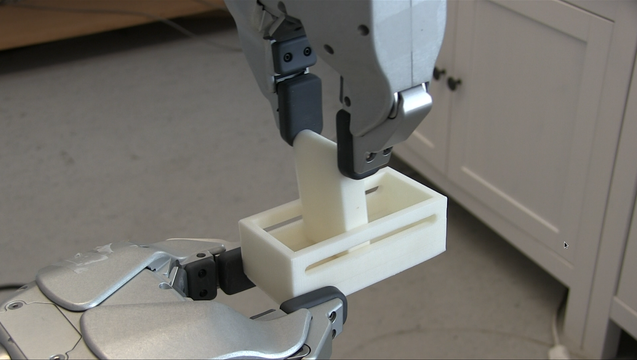}
		\caption{\label{vid2}}
	\end{subfigure}
	\begin{subfigure}[b]{0.13\textwidth}
		\includegraphics[width = \textwidth]{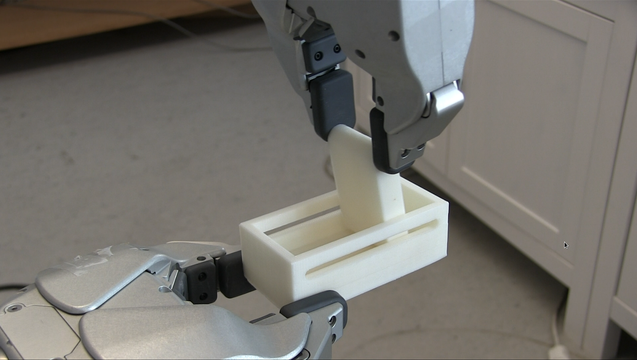}
		\caption{\label{vid3}}
	\end{subfigure}
	\\
	\begin{subfigure}[b]{0.13\textwidth}
		\includegraphics[width = \textwidth]{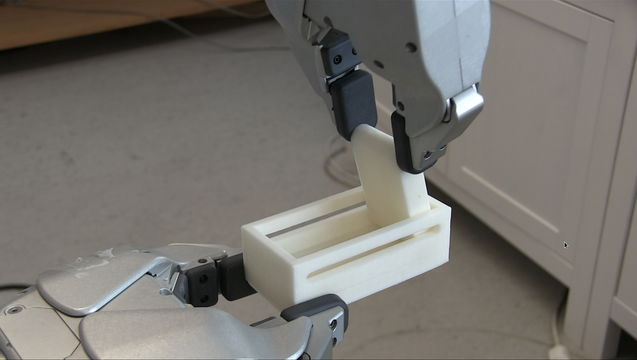}
		\caption{\label{vid4}}
	\end{subfigure}
	\begin{subfigure}[b]{0.13\textwidth}
		\includegraphics[width = \textwidth]{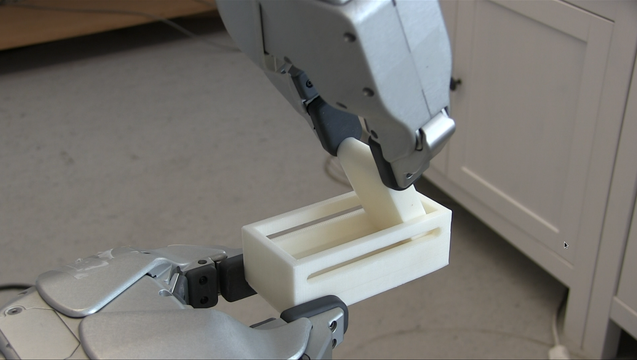}
		\caption{\label{vid5}}
	\end{subfigure}
	\begin{subfigure}[b]{0.13\textwidth}
		\includegraphics[width = \textwidth]{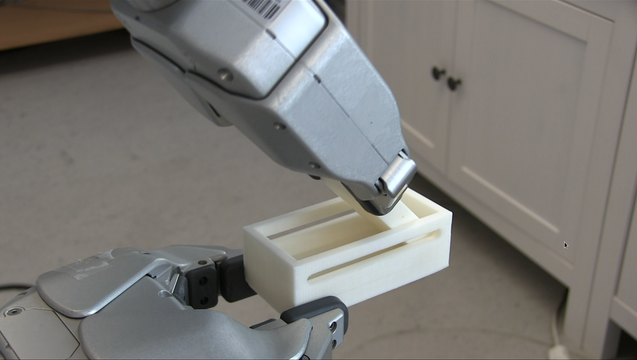}
		\caption{\label{vid6}}
	\end{subfigure}
	\caption{Folding assembly execution \label{assemblyRun}}
\end{figure}

The feedback linearisation control law \eqref{controlEquation} requires knowledge of the rod piece end-effector angular velocity and of the vector $\mathbf{r}_1 = \mathbf{p}_c - \mathbf{p}_1$.
We obtain $\mathbf{p}_1$ using forward kinematics and proprioception while the end-effector angular velocity is assumed to approximately equal to  the commanded value $\boldsymbol \omega_1$.
The contact point position $\mathbf{p}_c$ needs to be tracked  during task execution in order to correctly compensate for the term $\mathbf{S}(\boldsymbol \omega_1)(\mathbf{p}_c - \mathbf{p}_1)$.
Inexact compensation of this term yields unintended sliding or excessive contact forces.
A simple way to calculate $\mathbf{p_c}$ is by employing force torque sensors.

While both end-effectors will measure forces that are related to the contact point position, the rod piece end-effector measurements are distorted by inertial forces that makes the contact point computation challenging when the task is executed in high speed.
However following the master-slave approach, we can take advantage of the static surface piece end-effector and compute $\mathbf{r}_2$ by means of the relationship:
\begin{equation}
	\boldsymbol \tau_2 = \mathbf{S(r}_2) \mathbf{f}_2.
	\label{forceTorqueRelationship}
\end{equation}

\begin{remark}
It is important to note that, with a rigid grasp, the rod piece end-effector force torque measurements can be employed to estimate the contact point position with respect to the end-effector frame, as in \cite{Karayiannidis2014}.
\end{remark}

\begin{figure}
	\centering
	\includegraphics[width=0.45\textwidth]{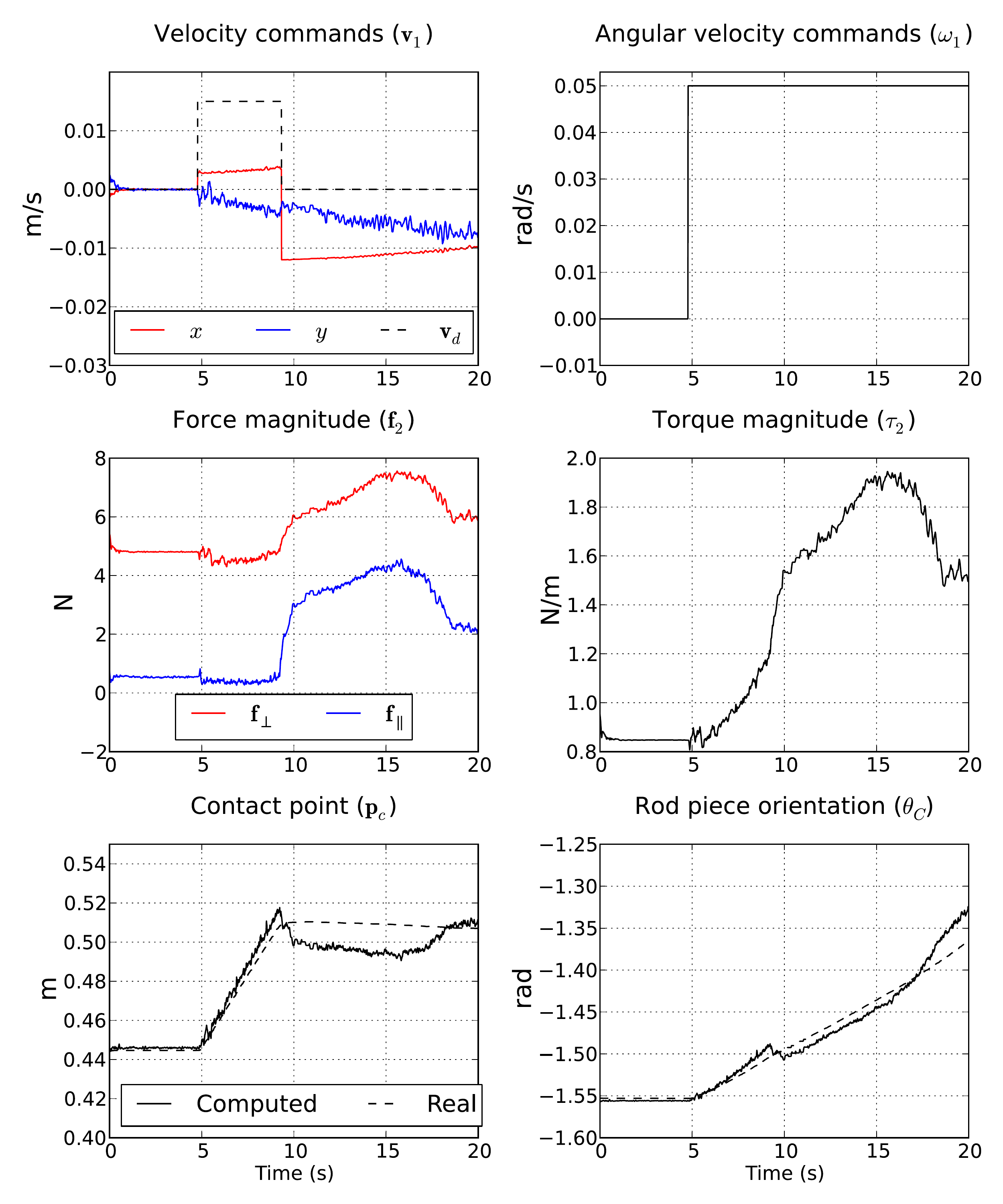}
	\caption{Command signals and contact point computation\label{assemblySignals}}
\end{figure}

\begin{figure}
	\centering
	\includegraphics[width=0.45\textwidth]{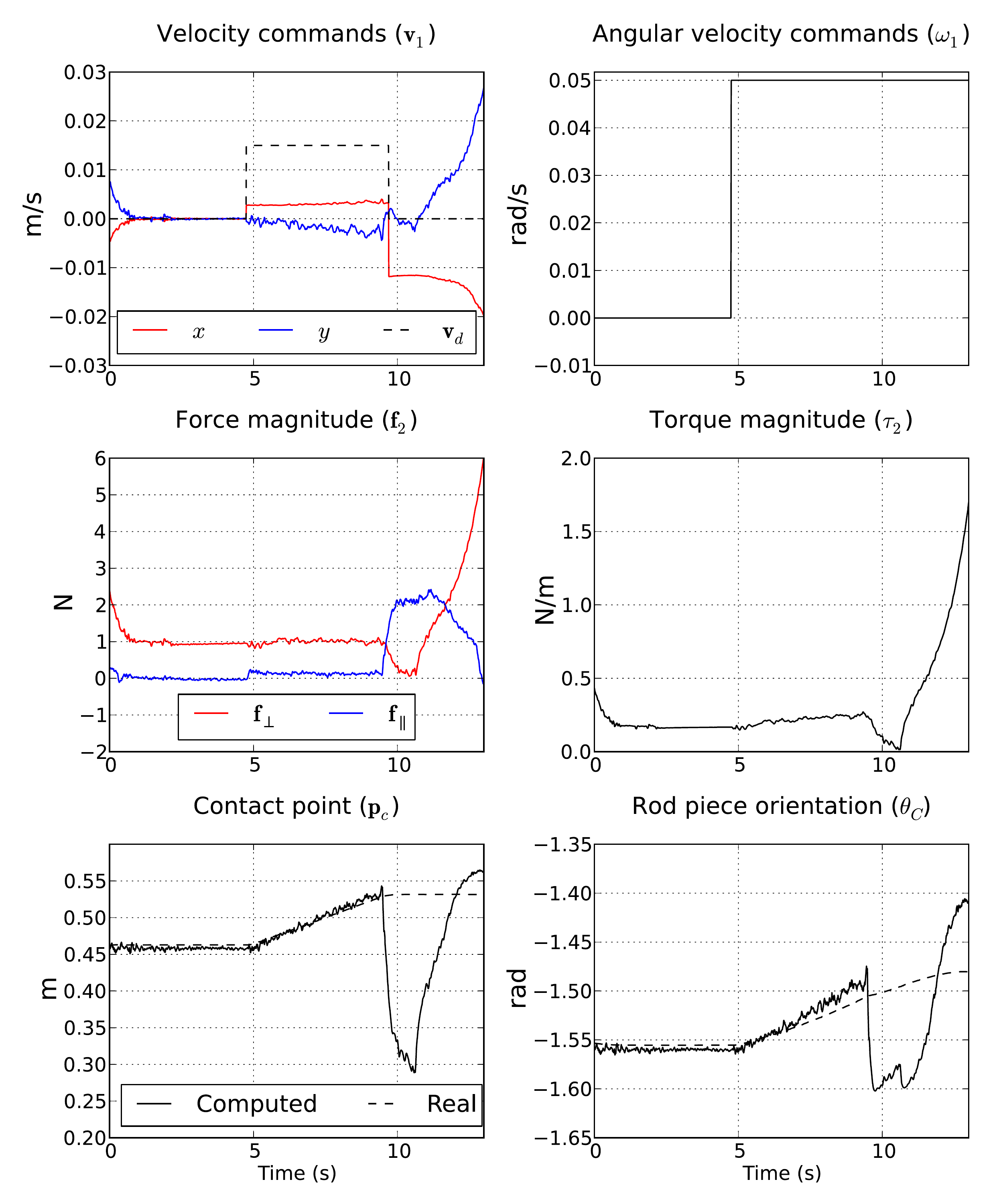}
	\caption{Values obtained when contact is lost\label{breakInContact}}
\end{figure}

Depending on the contact point being fixed or able to slide, the constraints on the system change and $\mathbf{f}_2$ may have components along the surface piece tangent.
By projecting the contact forces along the surface piece normal, we can compute $\mathbf{r}_2$ magnitude with
\begin{equation}
	\|\mathbf{r}_2\| = \frac{\|\boldsymbol \tau_2\|}{\|\mathbf{f}_2^\top \mathbf{n}\|}.
	\label{r2Computation}
\end{equation}
The orientation of $\mathbf{r}_2$ is chosen so as to comply with \eqref{forceTorqueRelationship}.
The contact point results trivially from the definition of $\mathbf{r}_2$ \eqref{rVectors}, and the orientation of the rod piece with respect to the surface piece can be parameterised as an $\theta_c$:

\begin{equation}
	\theta_c = \arctan{\frac{\mathbf{n}^\top \mathbf{r}_1 \|\mathbf{r}_2\|}{\mathbf{r}_2^\top \mathbf{r}_1} },
\end{equation}
defined with respect to the same direction as the measured torques.

\section{Experimental setup and results\label{results}}
We experimentally tested the folding assembly execution with a PR2 dual-arm manipulator from Willow Garage, equipped with wrist force torque sensors.
For the experiments, we apply the control input (\ref{controlEquation}) in order to execute a desired contact point motion without breaking contact and  to showcase the viability of folding.  To do so, we designed a pair of objects to work as the surface and rod pieces (fig. \ref{foldingObject}).

The experiment is initialized by having the robot move its end-effectors to pre set locations and manually feeding it the parts, ensuring initial contact conditions (fig. \ref{vid1}).
It then uses its force measurements to get the direction of the surface normal.

We used $K_f = 0.01$, $f_d = 5$N and $\mathbf{f}_d = f_d \mathbf{n}$.
The force feedback gain $K_f$ is set so that, for expected errors in the order of magnitude of Newtons, the absolute value of $\mathbf{v}_f$  remains in the order of cm$/$s.
The desired contact point velocities absolute values were set as $v_d = 0.015$ m$/$s, $\omega_{\text{ref}} = \omega_d = 0.05$ rad$/$s until the rod piece hits the wall at the end of the surface piece and the contact point becomes fixed.
We then use $v_d = 0$ m$/$s and maintain the angular velocity.
Furthermore, an average of the force torque sensor values over the previous five samples was taken before feeding them to the controller, to mitigate the effects of sensor noise.
The controller was run at $100$Hz, with the inner velocity control loop at $1000$Hz.
Results are shown in fig. \ref{assemblySignals}, with frames from the observed motion in fig. \ref{assemblyRun}.

The robot was able to perform the intended assembly movement in a smooth way.
However, in addition to sensor noise, two drawbacks of relying solely on the force torque sensor measurements to obtain the contact point position and orientation of the rod piece are patent: a) when the magnitude of the contact forces increases by a significant margin, the contact point calculation suffers from a larger error with respect to the ground truth.
Imperfect definition of the surface piece normal yields to small tangential forces that affect the contact point derivation.
b) On the other hand, defining a smaller desired contact force can easily lead to a break in contact, from which this simple strategy cannot recover (fig. \ref{breakInContact}).

The dual-arm setup allows for force torque sensing on both hands.
Sensory information should be exploited fully in order to robustly estimate the contact point and surface normal, preventing the aforementioned issues.
Additionally, more advanced force control inputs  $\mathbf{v}_f$ could be designed in order to prevent large changes in the contact force while not breaking the unilateral constraint.

\section{Conclusion\label{conclusion}}
This paper proposed folding assembly as an assembly primitive that allows for two pieces to be assembled by means of a translational plus rotational movement.
We modelled the two part system as a passive-revolute joint located at the contact point.
The effects of using different grasps were considered, from a kineto-statics perspective.
Furthermore, we proposed a simple controller based on the direct contact point calculation through force torque measurements.
The viability of folding was demonstrated by the successful implementation of the proposed controller.
Merged with robust estimation of the contact point, we believe that folding can be easily integrated in a higher level assembly planning and execution algorithm, increasing the available toolset for approaching complex assembly tasks.
\bibliographystyle{unsrt}
\bibliography{biblio}
\end{document}